\documentclass[runningheads]{llncs}
\usepackage[T1]{fontenc}
\usepackage{graphicx}
\usepackage{amsmath}
\begin{document}
\title{Uncertainty-Aware 3D Residual Wavelet Diffusion for Ultra Low-Field MRI Super-Resolution}
\titlerunning{3D Residual Wavelet Diffusion for Low-Field MRI SR}
\author{%
Rui W. Yeow \and
Millie Beament \and
Fred Dick \and
Raha Razin \and
Martina Bocchetta \and
David L. Thomas \and
Henry F. J. Tregidgo \and
Daniel C. Alexander \and
James H. Cole
}
\authorrunning{R. W. Yeow et al.}
\institute{University College London, London, UK \\
    \email{rui.yeow.24@ucl.ac.uk}}
\maketitle
\begin{abstract}
Ultra low-field MRI expands global access to neuroimaging but produces scans with low signal-to-noise ratio, reduced contrast, and thick slices. While regression-based super-resolution can recover anatomical detail for segmentation, it returns a single deterministic estimate that gives no indication of regions where the low-field input leaves anatomy underdetermined. Generative diffusion models offer an alternative by sampling the posterior distribution of plausible high-field images, quantifying this anatomical ambiguity. However, applying them to 3D whole-brain MRI is restricted by memory bottlenecks, slow sampling, and scanner domain shifts. We propose a 3D residual wavelet diffusion model that combines three ideas to overcome these hurdles. A lossless wavelet reparameterisation shrinks the spatial grid to fit a whole brain on a single GPU, residual shifting accelerates sampling by starting from the low-field input, and domain randomisation promotes scanner generalisation without paired training data. As the high-field reference is not a voxel-aligned ground truth, we evaluate downstream volumetric agreement. On a healthy cohort ($n=19$) imaged at 0.064\,T and 3\,T, our method matches a leading general-purpose regression approach in volumetric accuracy while additionally generating per-voxel uncertainty maps highlighting underdetermined regions. Furthermore, on a pilot dataset ($n=11$) of participants with cognitive impairment, disease-relevant atrophy is preserved rather than normalised towards a healthy prior. Our framework brings whole-brain posterior sampling to low-field super-resolution without sacrificing volumetric accuracy.

\end{abstract}
\section{Introduction}

Magnetic resonance imaging (MRI) remains out of reach for much of the world, with Japan operating around 50 scanners per million people and West Africa fewer than one~\cite{ogbole2018survey}. This disparity reflects more than the purchase price of high-field scanners, as they also demand reliable power, radiofrequency (RF) shielding, and continual maintenance, with donated systems often ending up in MRI graveyards in low-resource settings~\cite{jones2025low}. Ultra low-field scanners (e.g., 0.064\,T) offer a promising alternative, requiring no RF shielding and minimal power. However, the low field strength limits image quality, with low signal-to-noise ratios (SNR), reduced tissue contrast, and thick slices~\cite{marques2019low}. Consequently, the fine anatomical detail required for quantitative measurement is often unavailable and must be recovered~\cite{iglesias2022quantitative}.

Super-resolution can recover this detail, synthesising high-field images that enable downstream tasks such as morphometry and volumetric segmentation~\cite{iglesias2022quantitative,sorby2024portable}. One common approach uses regression-based models trained with domain randomisation on synthetic data (e.g., SuperSynth~\cite{liu2025modality} and LF-SynthSR~\cite{sorby2024portable}). On real 0.064\,T scans, these models recover volumes correlating with high-field references~\cite{iglesias2022quantitative} and can separate Alzheimer's disease (AD) and mild cognitive impairment (MCI) individuals from controls~\cite{sorby2024portable}.

Yet because super-resolution is an ill-posed inverse problem, a single low-field scan is consistent with many different high-field images. A regression model trained to minimise reconstruction error returns the posterior mean. While this achieves the lowest average error, the averaging process produces a smooth output less detailed than real high-field brains~\cite{blau2018perception}. More importantly, a deterministic estimate provides no information about the spread of plausible anatomies, failing to distinguish between areas dictated by the low-field signal and where details are underdetermined. Conversely, a conditional diffusion model approximates the whole posterior. It learns a score function to sample high-field volumes consistent with the low-field input~\cite{song2020score}. Drawing multiple samples gives a set of plausible reconstructions, from which we can compute a posterior mean and a per-voxel map of sample disagreement that deterministic regression cannot provide~\cite{luo2023bayesian}.

Extending this posterior distribution to a full 3D low-field brain presents three challenges: computational constraints limit 3D diffusion to 2D slices or downsampled volumes~\cite{friedrich2024wdm}; traditional sampling is slow~\cite{safari2025mri}; and models trained on one scanner often fail to generalise to others, made worse by the scarcity of paired training data~\cite{islam2023improving}. We propose a 3D residual wavelet diffusion model to overcome these challenges. A lossless wavelet decomposition shrinks the spatial grid eightfold to fit whole brains on a single GPU without discarding fine detail~\cite{friedrich2024wdm}. Residual diffusion begins sampling from the low-field input, reaching a reconstruction in tens of steps~\cite{yue2023resshift}. Domain randomisation trains on synthetic data spanning broader contrast and resolution ranges than any single scanner, enabling unpaired generalisation~\cite{billot2023robust}.
To our knowledge, no prior work has combined these approaches for 3D super-resolution. Doing so makes whole-brain posterior sampling feasible without sacrificing volumetric accuracy, exposing where the low-field input underdetermines the anatomy rather than returning a single, uniformly confident estimate. Evaluated on cohorts imaged at 0.064\,T and 3\,T, our method achieves volumetric agreement competitive with the regression benchmark and preserves disease-relevant signal.

\section{Method}

\begin{figure}[t]
\centering
\includegraphics[width=0.9\textwidth]{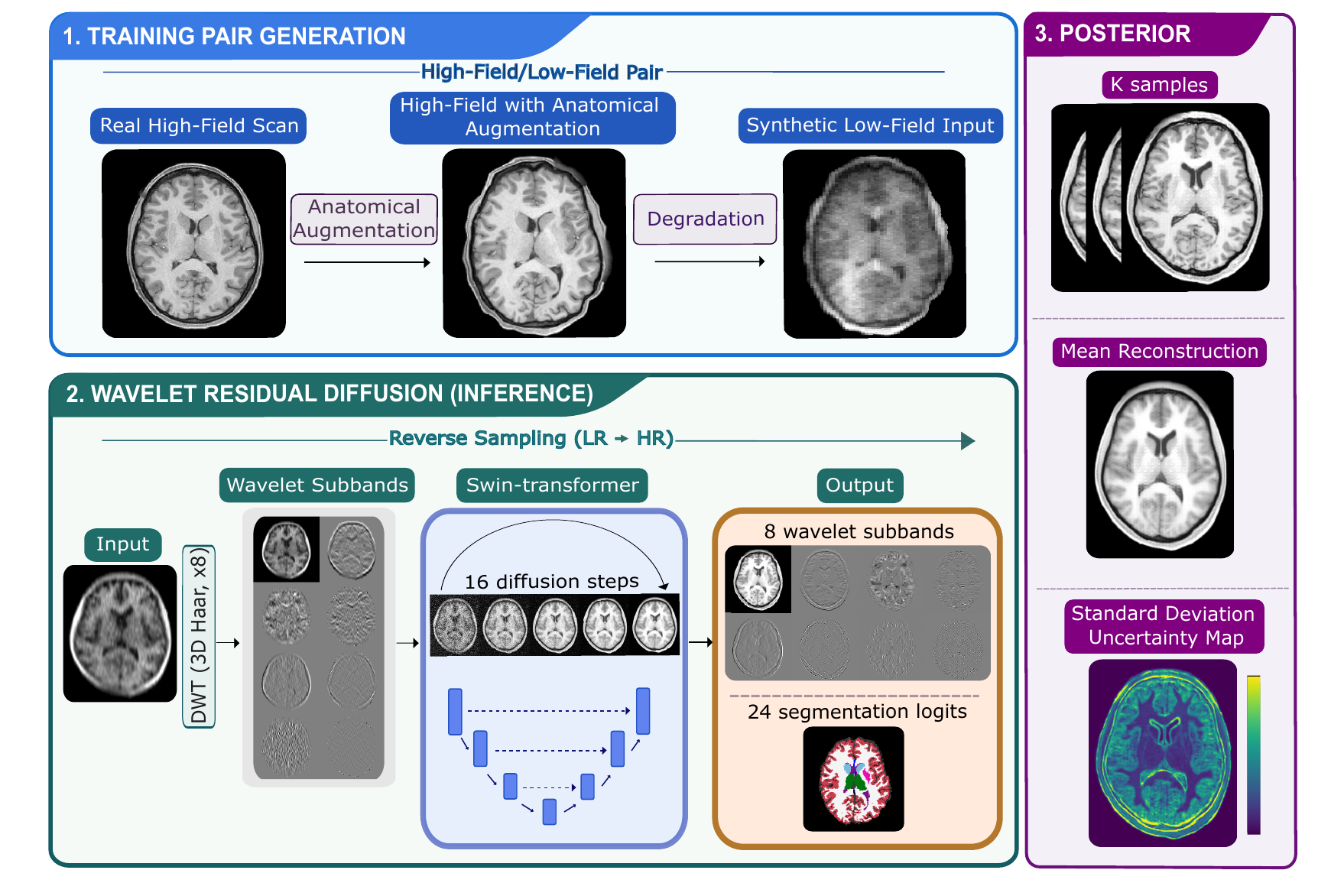}
\caption{Proposed pipeline. (1) Synthetic training pairs are formed by augmenting and degrading high-field scans. (2) A Swin transformer U-Net converts the low-field input to Haar subbands and super-resolves them via a residual-shift reverse process. (3) Drawing 
$K$ samples produces a mean reconstruction and per-voxel standard-deviation map.}
\label{fig:pipeline}
\end{figure}

We frame super-resolution as a conditional diffusion process that samples the posterior distribution of high-field volumes consistent with a low-field input, rather than regressing a single output (Fig.~\ref{fig:pipeline}). The model pairs a forward process, which interpolates a high-field volume towards its low-field counterpart while adding noise, with a reverse diffusion process that inverts it, both defined on the volumes' wavelet coefficients. A trained U-Net transformer conditioned on the low-field input drives the reverse process, and running it repeatedly produces a set of reconstructions whose per-voxel mean and standard deviation form the output.

\subsection{Wavelet-domain reparameterisation}

The memory used by a 3D diffusion network scales with the number of voxels it processes. A whole-brain volume has too many voxels to fit a 3D diffusion model on a single GPU, so we must reduce the number of spatial positions without discarding the fine anatomical detail that super-resolution has to recover. A 3D Haar wavelet transform does this losslessly, as in 3D wavelet diffusion~\cite{friedrich2024wdm}. It replaces every $2\times2\times2$ block of voxels with eight coefficients, one local average and seven differences. The averages across the volume form a half-resolution copy of the brain, and the differences hold its high-frequency detail. Stacking these eight subbands as channels leaves one-eighth the spatial positions to process, and roughly one-eighth the memory footprint. Because the transform is orthonormal, its inverse is the transpose and recovers the original volume exactly.

\subsection{Residual-shift diffusion}

Let $x_{\mathrm{HR}}$ and $x_{\mathrm{LR}}$ be the Haar coefficients of the high-field target and the low-field input on the half-resolution grid. The forward process follows the residual-shift construction~\cite{yue2023resshift}, interpolating from the high-field target towards the low-field input as the noise grows. Its marginal at time $t$ is
\begin{equation}
q(x_t \mid x_{\mathrm{HR}}, x_{\mathrm{LR}}) = \mathcal{N}\!\big( x_{\mathrm{HR}} + (x_{\mathrm{LR}} - x_{\mathrm{HR}})\beta_t,\ \gamma^2 \beta_t I \big),
\end{equation}
where $\beta_t$ increases monotonically with $t$ from near zero to near one and $\gamma$ scales the isotropic noise. At $t=0$ the distribution is concentrated on the high-field target, and at 
$t=1$ centred on the low-field input with maximum variance. The endpoint is a noisy low-field volume rather than pure noise~\cite{safari2025mri}.

To invert this process, a network $f_\theta$ predicts the clean target $\hat{x}_0$ from the current state $x_t$, the low-field conditioning $x_{\mathrm{LR}}$, and the time $t$. We adopt $x_0$ parameterisation rather than predicting noise because it keeps the regression target on the scale of the data and is equivalent to learning the score of the diffusion process~\cite{song2020score}. Sampling begins from a noisy version of the low-field input rather than pure Gaussian noise. Specifically, we draw the initial state $x_1 \sim \mathcal{N}(x_{\mathrm{LR}}, \gamma^2\beta_{\max}I)$ at $t=1$, which relies solely on the low-field input. At each step, the network predicts the target and re-noises it to the next earlier time $t'$ following the same noise schedule~\cite{yue2023resshift}.
\begin{equation}
\hat{x}_0 = f_\theta(x_t, x_{\mathrm{LR}}, t), \qquad
x_{t'} = \hat{x}_0 + (x_{\mathrm{LR}} - \hat{x}_0)\beta_{t'} + \gamma\sqrt{\beta_{t'}}\,z, \quad z \sim \mathcal{N}(0, I),
\end{equation}
where $x_{\mathrm{LR}}$ conditions the network at every step and $z$ is fresh noise. Iterating from $t=1$ to $t=0$ over a few steps draws one sample from the posterior of high-field volumes consistent with the input.

\subsection{Network and conditioning}\label{sec:network}

Each reverse step is computed by a 3D Swin transformer U-Net~\cite{liu2021swin}, whose windowed self-attention scales linearly with the number of voxels rather than the quadratic cost of global attention. We encode relative position with 3D rotary embeddings~\cite{su2024roformer} and condition the network on the low-field input by concatenating its eight subbands with the eight noisy-state subbands into a sixteen-channel input. The final U-Net layer uses a shared head to predict both the wavelet subbands and an auxiliary segmentation map. The segmentation branch avoids a separate high-resolution decoder by predicting the eight voxels of each $2\times2\times2$ block at half-resolution, which a pixel shuffle operation reassembles to full resolution. During training, this segmentation is supervised against FreeSurfer's automatic subcortical segmentation (\texttt{aseg})~\cite{fischl2002whole} (October 2025 build). This auxiliary task forces the shared feature map to encode morphology, anchoring the reconstruction to anatomical structure. We discard the segmentation output at inference~\cite{laso2024quantifying}.
\subsection{Domain-randomised training pairs}

Each training pair is synthesised on the fly from a real high-field volume to share identical anatomy but differ purely in acquisition. First, we augment the high-field scan anatomy (random affine transforms, elastic deformations, flips) to broaden the training set without simulating atrophy. Second, we degrade this volume into a simulated low-field input. We randomise tissue contrast by softly grouping voxels into three intensity-based classes --- grey matter, white matter, and cerebrospinal fluid (CSF) --- then remapping each tissue to a target value sampled from a wide range. Each voxel keeps its original deviation from the tissue mean, preserving real intra-tissue variation. We then apply a multiplicative bias field to model RF coil inhomogeneity, generated by sampling a coarse $4\times4\times4$ Gaussian grid, upsampling it smoothly, and exponentiating it. We then apply motion blur. To model resolution loss, we operate in k-space~\cite{lin2023low}, retaining central Fourier coefficients up to a random resolution, injecting Rician noise for low SNR, and resampling via nearest-neighbour interpolation. Randomising contrast and resolution far beyond any single device trains the model to be robust across acquisitions rather than tuned to one scanner~\cite{billot2023robust}.

\subsection{Training objective}

The objective is a sum of four terms,
\begin{equation}
\mathcal{L} = \mathcal{L}_{\mathrm{img}} + \mathcal{L}_{\mathrm{wav}} + \mathcal{L}_{\mathrm{LPIPS}} + \lambda\,\mathcal{L}_{\mathrm{seg}}.
\end{equation}
$\mathcal{L}_{\mathrm{img}}$ is an $L_1$ loss between the reconstructed image and target volumes, after the inverse transform~\cite{durrer2025fastwdm3d}. $\mathcal{L}_{\mathrm{wav}}$ is an $L_1$ loss on the wavelet coefficients after normalising each subband to unit variance to prevent low-frequency dominance and push high-frequency recovery~\cite{guth2022wavelet}. $\mathcal{L}_{\mathrm{LPIPS}}$ applies Learned Perceptual Image Patch Similarity to the central slice along each axis, which sharpens detail the $L_1$ terms tend to smooth~\cite{zhang2018unreasonable}, and $\mathcal{L}_{\mathrm{seg}}$ supervises the segmentation output of the shared head using Dice and
cross-entropy~\cite{milletari2016v}.

\subsection{Inference}

At inference we draw $K$ independent reconstructions. Each starts from a terminal sample, is refined over the reverse process, and is mapped to voxel space by the inverse Haar transform, producing $x^{(k)}$. The super-resolution estimate is the per-voxel mean $\bar{x}$ (the minimum mean-squared-error reconstruction), and the uncertainty map $u$ is their per-voxel standard deviation,
\begin{equation}
\bar{x} = \frac{1}{K}\sum_{k=1}^{K} x^{(k)}, \qquad
u = \sqrt{\frac{1}{K-1}\sum_{k=1}^{K}\big(x^{(k)} - \bar{x}\big)^2}.
\end{equation}

\subsection{Implementation}

The Swin U-Net uses an embedding width of 48, four resolution levels with depths (2, 2, 6, 2) and heads (3, 6, 12, 24), and window size 8. The shared head produces 24 bilateral-merged FreeSurfer classes. We use a geometric schedule $\beta_t$ from $4 \times 10^{-4}$ to $0.9999$, discretising the interval $t \in [0,1]$ into 1000 steps for training (exponent 0.3, noise scale $\gamma = 2$), sampling with 16 steps and $K = 32$ posterior draws. Augmentation draws rotations $\pm10^\circ$, scaling [0.9, 1.1], shear [$-0.1, 0.1$], elastic deformation, and flips, degrading to resolutions as coarse as 6 mm with Rician noise. We optimise with AdamW~\cite{loshchilov2017decoupled} (lr $5 \times 10^{-5}$, warmup-cosine, bf16, batch size 1) and weight the segmentation loss by $\lambda = 0.1$. We train on 682 T1-weighted volumes from the Human Connectome Project~\cite{van2013wu} and 369 from the Alzheimer's Disease Neuroimaging Initiative~\cite{jack2008alzheimer}, all affine-registered to the MNI305 template at 1 mm, head-masked, intensity-normalised, and cropped to $216 \times 208 \times 232$.

\section{Experiments and Results}

Evaluating this framework requires departing from traditional image-restoration metrics. The model maps inputs to a high-field domain defined by its training data, so the paired 3\,T evaluation scan is not guaranteed to match that distribution. Under this expected distribution shift, no voxel-level ground truth exists. Our objective is accurate anatomical recovery, not a match to a specific scanner's contrast. Consequently, full-reference metrics such as PSNR and SSIM are inappropriate, as they would conflate scanner mismatch with reconstruction error. We therefore measure anatomical fidelity downstream, comparing the volume of each brain structure against the paired 3\,T reference. 

We evaluated two cohorts, both imaged at 0.064\,T and 3\,T. The low-field scans were acquired at $1.6\times1.6$\,mm in-plane with 5\,mm slices, and the 3\,T references at approximately 1\,mm. The first comprises 19 healthy volunteers (HV; 7 men, 12 women; aged 24--49), while the second includes 11 MCI/AD subjects from the cognitive disorders clinic (CDC) study (7 men, 4 women; aged 58--77; Mini-Mental State Examination scores 3--28)~\cite{rosa2025ultra}.

Measuring volumes requires an evaluator independent of the compared methods. We apply a single external tool, SynthSeg~\cite{billot2023robust}, to every output and the 3\,T reference. We choose SynthSeg because it is contrast-agnostic. This ensures it treats our synthesised high-field images and the actual 3\,T reference identically, preventing inter-scanner contrast differences from biasing the volumetric comparison. It also avoids biases from method-specific segmentations (e.g., SuperSynth's built-in tool). Prior to segmentation, all volumes are affinely registered to the MNI305 template with SynthMorph~\cite{hoffmann2021synthmorph}. Holding this pipeline fixed, the volumetric comparison reflects how well the low-field resolution is recovered.

To ensure a fair comparison, all methods are evaluated through this identical pipeline. Both regression baselines, like our method, are trained by domain randomisation on synthetic data rather than on paired scans. Agreement is summarised by Pearson's $r$, the consistency ICC(3,1) and the absolute-agreement ICC(2,1)~\cite{shrout1979intraclass}, the magnitude of the percentage volume difference from the reference, and Dice overlap. Differences are tested with a subject-level clustered permutation (Holm-corrected) with 95\% cluster-bootstrap confidence intervals~\cite{demvsar2006statistical}.

\paragraph{Volumetric agreement on the healthy cohort.}
We evaluate volumetric agreement on the HV cohort ($n=19$) across 18 bilateral-summed brain structures
spanning grey matter, white matter, ventricles, CSF, and the brainstem (Table~\ref{tab:agreement}). Cubic interpolation is a weak floor, agreeing poorly with the reference on every metric. Our diffusion posterior mean instead matches the regression benchmark, showing no significant difference from SuperSynth~\cite{liu2025modality} on any metric, but achieving significantly higher Dice overlap and lower volumetric bias than LF-SynthSR~\cite{sorby2024portable}.

\begin{table}[t]
\centering
\caption{Volumetric agreement with 3\,T on the healthy volunteers ($n=19$) across 18 bilateral-summed structures; medians with 95\% cluster-bootstrap confidence intervals in brackets. $^{*}$ significant vs.\ Ours (clustered permutation, Holm-corrected, $p<0.05$).}
\label{tab:agreement}
\small
\setlength{\tabcolsep}{4pt}
\begin{tabular}{lcccc}
\hline
 & Cubic interpolation & LF-SynthSR & SuperSynth & Ours \\
\hline
Pearson $r$ & 0.15 [$-$0.08, 0.47] & 0.73 [0.66, 0.82] & 0.73 [0.69, 0.87] & 0.71 [0.64, 0.81] \\
ICC(3,1) & 0.08$^{*}$ [$-$0.04, 0.30] & 0.72 [0.61, 0.77] & 0.72 [0.66, 0.84] & 0.70 [0.61, 0.79] \\
ICC(2,1) & 0.04$^{*}$ [$-$0.01, 0.10] & 0.51 [0.33, 0.59] & 0.60 [0.49, 0.68] & 0.47 [0.36, 0.60] \\
$|$bias$|$\,\% & 43.1$^{*}$ [32.2, 58.9] & 6.8$^{*}$ [4.8, 9.1] & 3.6 [2.7, 4.9] & 4.7 [4.1, 6.2] \\
Dice & 0.42$^{*}$ [0.32, 0.50] & 0.74$^{*}$ [0.72, 0.75] & 0.80 [0.76, 0.82] & 0.81 [0.81, 0.82] \\
\hline
\end{tabular}
\end{table}

\begin{figure}[t]
\centering
\includegraphics[width=0.9\textwidth]{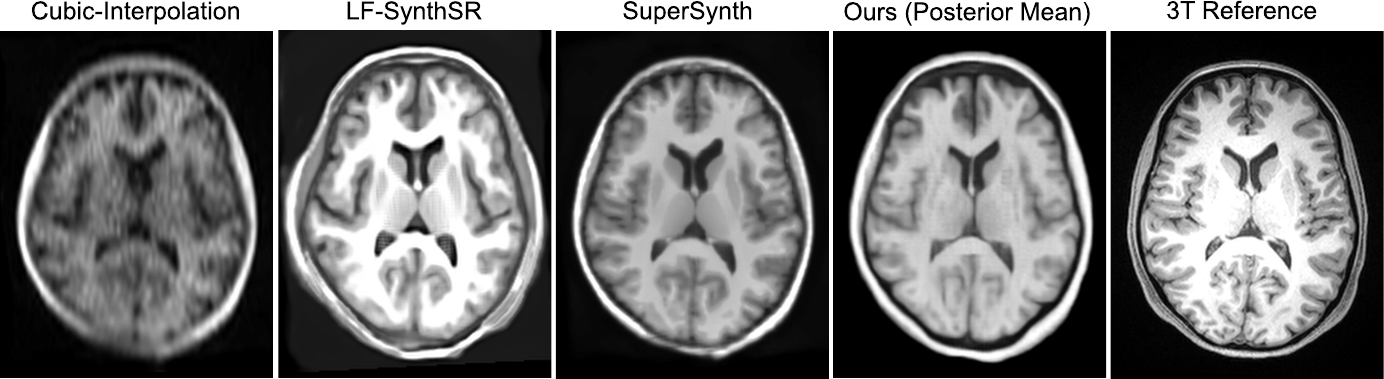}
\caption{Super-resolution of a 0.064\,T subject. Left to right: cubic interpolation of the low-field input, LF-SynthSR, SuperSynth, our posterior mean, and the 3\,T reference.}
\label{fig:montage}
\end{figure}

\begin{figure}[t]
\centering
\includegraphics[width=0.9\textwidth]{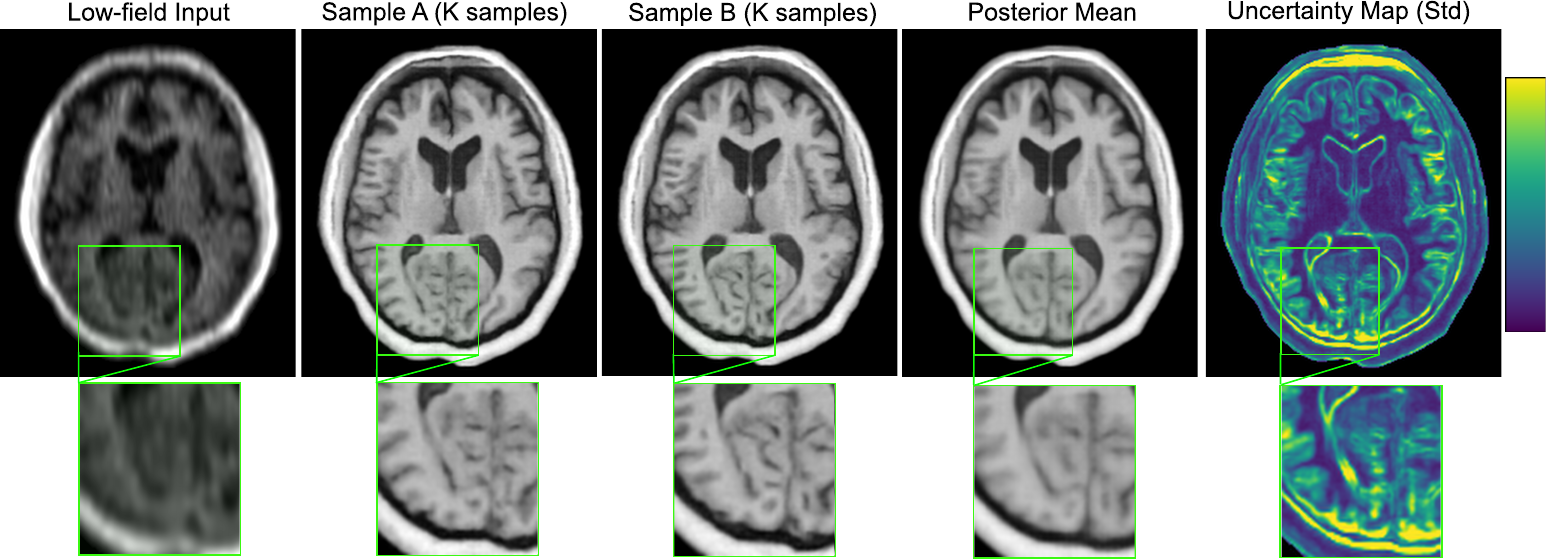}
\caption{Posterior for a subject imaged at 0.064\,T, on an axial slice. Samples A and B are sharp, plausible reconstructions, while the mean over thirty-two draws is smoother.}
\label{fig:posterior}
\end{figure}

Because the method matches the regression benchmark on these point estimates, the posterior is an addition to the reconstruction rather than a trade-off. The posterior mean recovers high-field-like contrast and anatomy (Fig.~\ref{fig:montage}), and although smoother than any single draw, remains the most accurate reconstruction on average. The samples behind it, each sharp and plausible, reveal where that reconstruction is underdetermined, disagreeing across the cortical folds, where the 0.064\,T input lacks the signal to resolve the anatomy and where cortical folding itself varies most between individuals (Fig.~\ref{fig:posterior}). The per-voxel standard deviation across the draws gives an uncertainty map that marks these underdetermined regions. Together the samples and the standard-deviation map carry information about where the reconstruction is underdetermined that a single deterministic output cannot provide~\cite{luo2023bayesian}. Importantly, the sharpness of these individual draws demonstrates that the smoothness of the posterior mean reflects anatomical ambiguity rather than a failure to generate high-frequency detail. By exposing multiple explanations for the same low-field input, the samples convey this ambiguity more intuitively than the uncertainty map alone.

\paragraph{Preservation of disease pathology.}
In the CDC dataset ($n=11$) of participants with MCI/AD, we tested whether super-resolution preserves hippocampal shrinkage and ventricular enlargement~\cite{barnes2009meta}. For each structure, we measured the correlation of volumes between the super-resolved scan and the paired reference across the cohort. The super-resolved lateral ventricle volumes track the reference at a Pearson $r$ of 0.81 [0.69, 0.97], and the hippocampus at 0.73 [0.25, 0.96]. While preliminary given the small sample, this indicates disease-related atrophy is preserved.

\section{Discussion and Conclusion}
We presented a 3D residual wavelet diffusion model that makes whole-brain posterior sampling feasible on a single GPU. By combining wavelet reparameterisation, residual diffusion, and domain randomisation, the model generalises to unseen low-field scanners in healthy and patient cohorts without paired data. On the healthy cohort, it is statistically indistinguishable from a leading general-purpose regression approach and is closer to the reference than the low-field-specialised baseline in segmentation overlap.

The uncertainty map accompanying each reconstruction is the method's principal advantage, though interpretable in only one direction. High standard deviations identify poorly constrained regions, but low values do not guarantee resolved anatomy, as low-variance structures produce similar samples regardless. Distinguishing these requires evaluating against natural population variance. Furthermore, although the small CDC dataset ($n=11$) limits definitive clinical conclusions, our preliminary results indicate that disease-related atrophy is preserved, motivating a properly powered future study to establish clinical utility. Even so, the method brings whole-brain posterior sampling to low-field super-resolution, matching the volumetric accuracy of the regression benchmark.

\begin{credits}
\subsubsection{\ackname}
This research was funded by the Alzheimer's Society Heather Corrie Impact Fund (grant number 577 [AS-PG-21-045]), Biogen Idec UK, the National Institute for Health and Care Research University College London Hospitals Biomedical Research Centre (NIHR UCLH BRC), and the Rosetrees Trust (CF-2022-2\textbackslash{}128).
\end{credits}
%
\bibliographystyle{splncs04}
\bibliography{references_new}
\end{document}